\pdfoutput=1

\documentclass[11pt]{article}

\usepackage[final]{acl}

\usepackage{times}
\usepackage{latexsym}
\usepackage[T1]{fontenc}
\usepackage[utf8]{inputenc}
\usepackage{microtype}
\usepackage{inconsolata}
\usepackage{url}
\usepackage{xspace}

\usepackage{amsmath}
\usepackage{amssymb}
\usepackage{mathtools}
\usepackage{textcomp}
\usepackage{siunitx}

\usepackage{graphicx}
\usepackage{graphbox}
\usepackage{booktabs}
\usepackage{multirow}
\usepackage{tabularx}
\usepackage{makecell}
\usepackage{subcaption}
\usepackage{tablefootnote}
\usepackage{float}
\usepackage[table]{xcolor}
\usepackage{colortbl}

\usepackage{enumitem}
\usepackage[hang]{footmisc}
\usepackage[normalem]{ulem}
\usepackage{soul}
\usepackage{fontawesome}
\usepackage[most]{tcolorbox}
\usepackage{pgf-pie}

\usepackage{tikz}
\usepackage{pgfplots}
\usetikzlibrary{shapes, arrows, positioning}
\pgfplotsset{compat=1.18}


\tikzstyle{arrow} = [thick, ->, >=stealth]
\tikzstyle{process} = [rectangle, minimum width=1.5cm, minimum height=0.7cm, text centered, draw=black, fill=yellow!5]

\definecolor{tikzblue}{RGB}{176,224,230}
\definecolor{tikzgreen}{RGB}{220,251,152}
\definecolor{bblue}{HTML}{4F81BD}
\definecolor{rred}{HTML}{C0504D}
\definecolor{ggreen}{HTML}{9BBB59}
\definecolor{ppurple}{HTML}{9F4C7C}
\definecolor{firebrick}{RGB}{178,34,34}
\definecolor{darkblue}{rgb}{0,0,0.5}

\hypersetup{
  colorlinks=true,
  citecolor=darkblue,
  linkcolor=darkblue,
  urlcolor=darkblue
}

\newcommand{\cc}[1]{%
  \pgfmathsetmacro{\mymix}{#1 * 0.8}%
  \edef\mycolor{\noexpand\cellcolor{firebrick!\mymix}}%
  \mycolor #1%
}

\newcommand{\ccn}[1]{%
  \pgfmathsetmacro{\val}{max(min(#1,100),-100)}%
  \pgfmathsetmacro{\mymix}{abs(\val)}%
  \ifnum#1<0
    \edef\mycolor{\noexpand\cellcolor{blue!\mymix}}%
  \else
    \edef\mycolor{\noexpand\cellcolor{red!\mymix}}%
  \fi
  \mycolor #1%
}

\newcommand{\ours}{\textsc{LegalScope}\xspace}
\newcommand{\nmodel}{28\xspace}
\newcommand{\nexam}{861\xspace}
\newcommand{\nrealcase}{276\xspace}
\newcommand{\njudgment}{56\xspace}
\newcommand{\nissue}{138\xspace}

\newcommand{\nrealcaseans}{7,728\xspace}
\newcommand{\nexamval}{80\xspace}
\newcommand{\nrealcaseval}{10\xspace}
\newcommand{\nexamvalans}{2,240\xspace}
\newcommand{\nrealcasevalans}{280\xspace}
\newcommand{\ntopicmin}{20\xspace}
\newcommand{\nexamus}{603\xspace}
\newcommand{\nexamcn}{94\xspace}
\newcommand{\nexamuk}{86\xspace}
\newcommand{\nexamau}{78\xspace}
\newcommand{\nexamen}{767\xspace}

\newcommand{\rqone}{How do current LLMs perform across the two tracks of \ours, and how well does exam performance transfer to real cases?}
\newcommand{\rqtwo}{Where and how does real-case legal reasoning fail across models and case types?}
\newcommand{\rqthree}{Can model-judge evaluation reliably approximate expert legal review?}

\newcommand{\cmark}{\checkmark}
\newcommand{\pmark}{$\triangle$}
\newcommand{\nmark}{-}
\newcommand{\gmark}{\textcolor{green!50!black}{\checkmark}}

\title{\ours: Measuring Exam-to-Case Transfer in\\ LLM Legal Reasoning}

\author{
  \textbf{Hongyu Wang\textsuperscript{1}}\quad
  \textbf{Rilyn R. Han\textsuperscript{2}}\quad
  \textbf{Yilun Zhao\textsuperscript{2}} \quad
  \textbf{Xuandong Zhao\textsuperscript{3}}\quad
  \textbf{Arman Cohan\textsuperscript{2}} \\[0.5em]
  \textsuperscript{1}UC Santa Barbara\quad
  \textsuperscript{2}Yale University\quad
  \textsuperscript{3}UC Berkeley \\
  \small{Corresponding to: \href{mailto:yilun.zhao@yale.edu}{yilun.zhao@yale.edu}}
}

\begin{document}
\maketitle

\begin{abstract}

Public legal exams are widely used to evaluate legal reasoning in large language models, yet it remains unclear whether exam performance reflects the ability to reason over real-case facts.
We introduce \ours, a dual-track benchmark that pairs \nexam legal-exam questions with \nrealcase lawyer-reviewed issue-stance prompts derived from de-identified Chinese judgments.
\ours evaluates models under both reference-answer scoring and case-based rubric scoring, enabling direct analysis of exam-to-case transfer.
Across \nmodel model groups, public-exam scores correlate with real-case scores (Pearson $r=0.817$) but do not fully predict them; citation relevance receives lower mean scores than argument validity under both automatic and lawyer evaluation.
We further show that automated evaluation aligns strongly with human review on exam answers but less reliably on real-case legal analysis, highlighting the need for expert-grounded case evaluation.

\end{abstract}

\begin{center}
\small
\setlength{\tabcolsep}{0pt}
\renewcommand{\arraystretch}{1.3}
\newcommand{\linkicon}[1]{\raisebox{-\dimexpr1.1\fontchardp\font`\g+0.1\fontcharht\font`\H\relax}{\includegraphics[height=1.2\dimexpr\fontcharht\font`\H+\fontchardp\font`\g\relax]{#1}}}
\begin{tabular}{c@{\hspace{5pt}}l@{\hspace{6pt}}l}
\linkicon{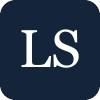} & \textbf{Homepage:} & \href{https://eternwang.github.io/LegalScope/}{\footnotesize\texttt{eternwang.github.io/LegalScope}} \\
\linkicon{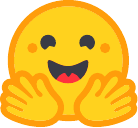} & \textbf{Data:} & \href{https://huggingface.co/datasets/Hongyu801/LegalScope}{\footnotesize\texttt{Hongyu801/LegalScope}} \\
\linkicon{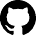} & \textbf{Code:} & \href{https://github.com/EternWang/LegalScope}{\footnotesize\texttt{EternWang/LegalScope}} \\
\end{tabular}
\end{center}

\section{Introduction}

LLMs are increasingly used and evaluated on legal tasks, from document review and question answering to bar-exam style problems~\cite{shi2026plawbenchrubricbasedbenchmarkevaluating, pires2026magisbenchevaluatingllmsmagistratelevel, chen2026legalcitebenchevaluatingcitationreliability, lee2026evaluatinglegalreasoningtraces}.
Standardized benchmarks such as LexGLUE~\cite{chalkidis2022lexgluebenchmarkdatasetlegal}, LegalBench~\cite{guha2023legalbench}, and LawBench~\cite{fei2024lawbench} track this progress at scale, and recent studies show that strong general-purpose LLMs reach high accuracy on professional exams and statutory reasoning tasks~\cite{DBLP:journals/corr/abs-2212-14402, DBLP:conf/icail/Blair-StanekHD23, DBLP:journals/ail/Martinez25}.
As a result, public legal exams have become a default proxy for evaluating legal reasoning ability.

\begin{figure}[t]
\centering
\includegraphics[width=\columnwidth]{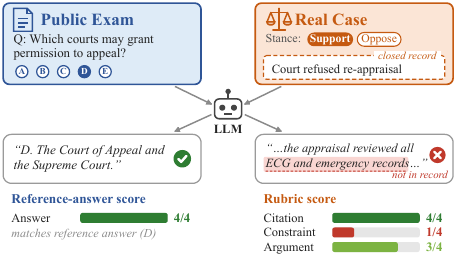}
\caption{Illustrative items from the two tracks of \ours. \textbf{Public Exam}: exam answers are scored against reference answers. \textbf{Real Case}: stance-specific analyses of de-identified Chinese judgments are scored on three rubric dimensions; adding facts beyond the closed record lowers constraint extraction.}
\label{fig:teaser}
\end{figure}

However, strong performance on public legal benchmarks does not necessarily indicate strong performance on real-case legal analysis.
Public exams are large-scale, standardized, and accompanied by reference answers, whereas real legal practice involves long fact patterns, incomplete information, contested interpretations, and competing lines of argument.
A useful legal response must not only recall relevant authorities, but also identify the operative conditions embedded in those authorities, connect those conditions to the facts under a controlled record, and construct a defensible argument under a specified stance.
Recent benchmarks have broadened this evaluation landscape through large-scale law-exam evaluation in LEXam~\citep{fan2026lexam}, user-centric practice-oriented tasks in UCL-Bench~\citep{gan-etal-2025-ucl}, and multiple-choice evaluation constructed from both recent legal cases and examination questions in LexGenius~\citep{liu-etal-2026-lexgenius}. However, the model-level relationship between independently scored public-exam and real-case tracks remains underexplored.

We introduce \ours, a dual-track benchmark designed to study this gap (\autoref{fig:teaser}).
The first track is a public-exam split of \nexam legal-exam instances drawn from the United States, China, the United Kingdom, and Australia.
The second track is a lawyer-reviewed real-case split of \nrealcase issue-stance prompts derived from \njudgment de-identified Chinese judgments and covering \nissue legal issues, with each issue paired into supporting and opposing prompts so that models must construct statute-grounded arguments under a specified stance rather than imitate the observed judgment outcome.
The exam track is scored by reference-answer matching, while the real-case track is scored under a calibrated rubric covering citation relevance, constraint extraction, and argument validity, with human validation on overlapping subsets.

Across \nmodel model groups, public-exam scores correlate with Chinese real-case scores at the model level (Pearson \(r=0.817\)), but rankings, reasoning-mode gains, and dimension-level strengths do not transfer uniformly across the two settings.
Citation relevance receives lower mean scores than argument validity under both automatic and lawyer evaluation (\S\ref{sec:analysis-validation}).
Automated evaluation aligns strongly with human review on exam answers (answer-level Pearson \(r=0.910\)) but weakens on real-case analysis when the two lawyers are pooled (\(r=0.312\)), indicating that scalable model-judge protocols do not yet match expert legal judgment in case-based settings.

Our contributions are fourfold.
\begin{itemize}[noitemsep,topsep=0pt,leftmargin=*]
    \item We introduce \ours, a dual-track benchmark that evaluates the same models on public legal exams and lawyer-reviewed real cases.
    \item We pair de-identified Chinese judgments into issue-stance prompts that test the same facts under supporting and opposing stances.
    \item We adapt a reference-aware exam scorer and a calibrated three-dimensional case rubric, both assessed against lawyer review.
    \item We find a strong but imperfect exam-to-case association, lower mean citation relevance than argument validity under both automatic and lawyer evaluation, and answer-level divergence between automatic and expert case scores.
\end{itemize}

\begin{figure*}[t]
\centering
\includegraphics[width=\textwidth]{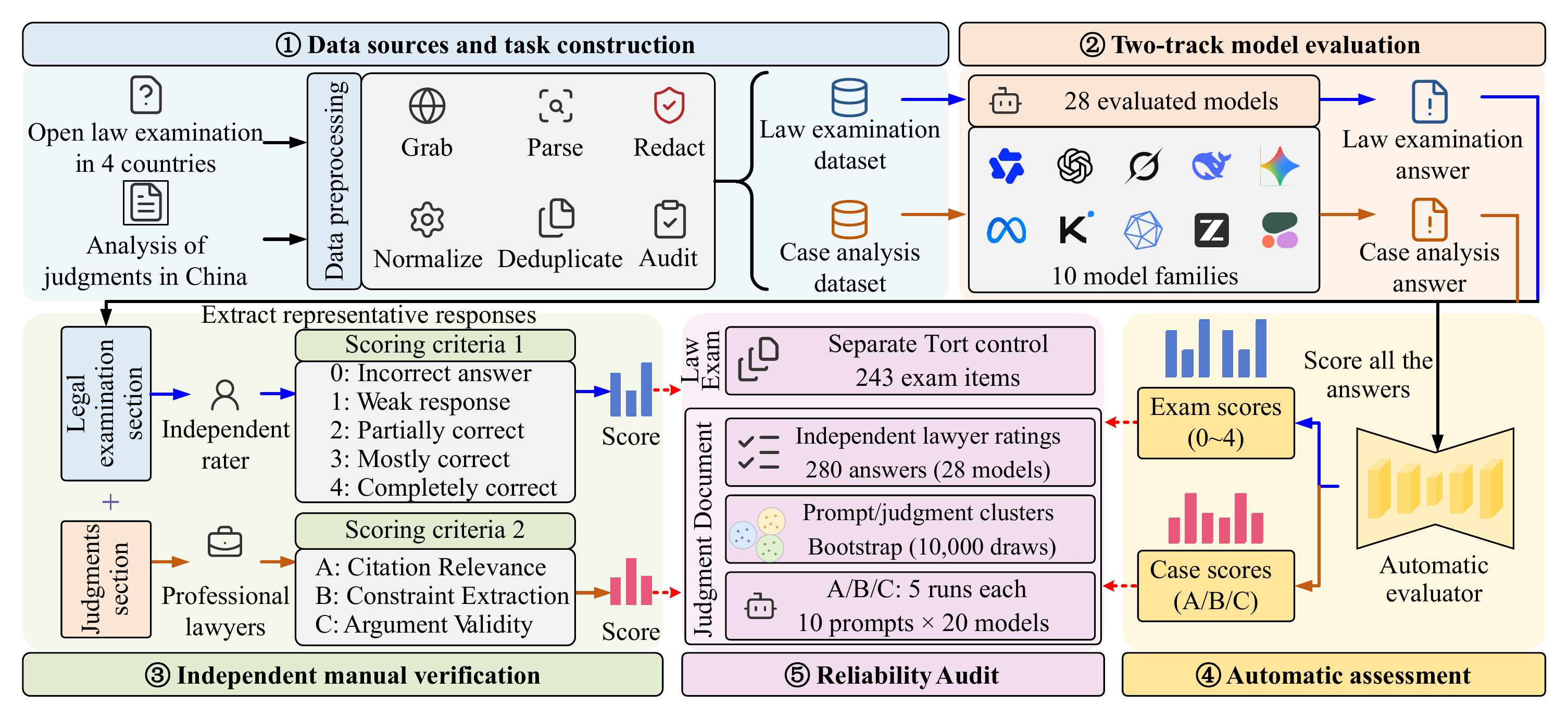}
\caption{Benchmark construction pipeline. Raw data are first collected from multiple sources, then de-identified and converted into structured records. QA pairs are then generated using carefully curated prompts. To ensure quality, human verification and annotation are applied during real-case curation and on validation subsets.}
\label{fig:benchmark-pipeline}
\end{figure*}

\section{Related Work}

\paragraph{Legal Benchmarks and Exam Proxies.}
Legal NLP evaluation has moved from task-specific datasets for judgment prediction, contract review, unfair-clause detection, and legal entailment~\cite{Aletras2016PredictingJD, xiao2018cail2018largescalelegaldataset, hendrycks2021cuad, koreeda2021contractnli, lippi2019claudette, wang2023maud} to broad suites such as LexGLUE and LegalBench, and, for Chinese law, JEC-QA, LeCaRDv2, LawBench, and LAiW~\cite{chalkidis2022lexgluebenchmarkdatasetlegal, guha2023legalbench, DBLP:conf/aaai/ZhongXTZ0S20, li2023lecardv2largescalechineselegal, fei2024lawbench, dai2025laiw}.
Public legal exams, standardized and paired with reference answers, have become a convenient proxy for LLMs, with widely debated bar-exam and statutory-reasoning results~\cite{DBLP:journals/corr/abs-2212-14402, DBLP:conf/icail/Blair-StanekHD23, DBLP:journals/ail/Martinez25} despite possible pretraining contamination~\cite{carlini2021extractingtrainingdatalarge, deng-etal-2024-investigating, xu2024benchmarkdatacontaminationlarge}.
Recent benchmarks such as LEXam, UCL-Bench, PLawBench, and LEGIT target open-ended and practice-oriented reasoning~\cite{fan2026lexam, gan-etal-2025-ucl, shi2026plawbenchrubricbasedbenchmarkevaluating, lee2026evaluatinglegalreasoningtraces}, and LexGenius mixes items from recent cases and exams~\cite{liu-etal-2026-lexgenius}.
As Table~\ref{tab:benchmark-comparison} shows, none of them tests the exam proxy against expert-reviewed real-case analysis when the same models are evaluated on both.

\begin{table}[t]
\centering
\small
\setlength{\tabcolsep}{4pt}
\renewcommand{\arraystretch}{1.05}
\resizebox{\columnwidth}{!}{%
\begin{tabular}{lcccccc}
\toprule
\textbf{Benchmark}
& \makecell{\textbf{Public}\\\textbf{exam}}
& \makecell{\textbf{Real}\\\textbf{cases}}
& \makecell{\textbf{Paired}\\\textbf{stance}}
& \makecell{\textbf{Expert-}\\\textbf{curated}\\\textbf{data}}
& \makecell{\textbf{Expert-}\\\textbf{validated}\\\textbf{scoring}}
& \makecell{\textbf{Exam-to-}\\\textbf{case transfer}} \\
\midrule
LexGLUE        & \nmark & \pmark & \nmark & \pmark & \nmark & \nmark \\
CUAD           & \nmark & \nmark & \nmark & \cmark & \nmark & \nmark \\
LegalBench     & \cmark & \nmark & \nmark & \cmark & \nmark & \nmark \\
LawBench       & \cmark & \pmark & \nmark & \pmark & \nmark & \nmark \\
CourtReasoner  & \nmark & \cmark & \cmark & \cmark & \cmark & \nmark \\
Bar Exam       & \cmark & \nmark & \nmark & \cmark & \pmark & \nmark \\
LEXam          & \cmark & \nmark & \nmark & \cmark & \cmark & \nmark \\
UCL-Bench      & \nmark & \pmark & \nmark & \cmark & \cmark & \nmark \\
LexGenius      & \cmark & \pmark & \nmark & \cmark & \nmark & \nmark \\
\midrule
\textbf{\ours} & \gmark & \gmark & \gmark & \gmark & \gmark & \gmark \\
\bottomrule
\end{tabular}%
}
\caption{Comparison of \ours with representative legal benchmarks. This table summarizes task coverage, not benchmark quality. \textbf{Expert-curated data}: domain experts author or review the benchmark materials themselves; \textbf{Expert-validated scoring}: automatic evaluation outputs are validated against expert judgment. $\checkmark$: full, $\triangle$: partial, \texttt{-}: no coverage.}
\label{tab:benchmark-comparison}
\end{table}

\paragraph{LLM Judges and Expert Review.}
Open-ended legal analysis cannot be scored by exact match.
LLM judges scale well but can disagree with human evaluators on literature-grounded tasks~\cite{zhao2025sciarena}.
Their judgments can reflect position and verbosity biases~\cite{ma-etal-2025-judging}, self-preference, and sensitivity to rubric wording~\cite{liu2023gevalnlgevaluationusing, kim2024prometheus2opensource}.
In legal answers, fluent prose can mask irrelevant or hallucinated citations~\cite{10.1093/jla/laae003, magesh2024hallucinationfreeassessingreliabilityleading}.
CourtReasoner evaluates U.S. judicial reasoning with paired stances and three expert-validated dimensions~\cite{han2025courtreasoner}.
We adapt this design to Chinese judgments, calibrate the rubric against lawyer comments, and measure where model judges and lawyers disagree.

\section{Benchmark Design}
\label{sec:benchmark-design}

\ours is organized into two complementary tracks designed to study how legal reasoning performance changes across evaluation settings.
The public-exam track is a scalable, reference-aware split that mirrors how prior work has measured legal ability on standardized exams.
The real-case track is a lawyer-reviewed, closed-book split that asks models to construct statute-grounded arguments over de-identified Chinese judgments.
\autoref{fig:benchmark-pipeline} illustrates the pipeline shared by both tracks.

\subsection{Public-Exam Split}
An instance in the public-exam split is one legal-exam question paired with a reference answer. Some source items retain answer choices and require a short explanation.
Each instance carries the question text, jurisdiction, legal category, reference answer, and a set of review constraints, such as restrictions on which authorities may be cited or which facts may be assumed.
A model answer is scored for substantive consistency with the reference rather than surface similarity, with the evaluator inspecting legal issues, governing rules, fact application, exceptions, and the final conclusion.
The split contains \nexam instances drawn from the United States (\nexamus), China (\nexamcn), the United Kingdom (\nexamuk), and Australia (\nexamau), after excluding one duplicate and six items with unverifiable shared context.
Sources span U.S.\ state bar exams along with NCBE and UBE materials, U.K.\ and Australian bar-equivalent exams, and Chinese legal-examination and training corpora; the full source breakdown appears in Appendix~\ref{app:source-domain-breakdowns}.

\subsection{Real-Case Split}
The real-case materials are concluded Chinese judgments reviewed by practicing lawyers after de-identification, covering seven legal categories; the original subset was contributed by a lawyer specializing in medical-damage disputes.
Throughout, ``real-case'' denotes provenance: prompts derive from real, concluded judgments and preserve their factual and evidentiary boundaries, rather than requiring reasoning over complete case files.
Each judgment is decomposed into the discrete legal issues addressed in its reasoning, producing \nissue issues over \njudgment judgments.
Every issue is rewritten as a closed-book factual background and paired into a supporting prompt and an opposing prompt, yielding \nrealcase issue-stance prompts in total.
For each prompt, the model is required to produce two paragraphs: one summarizing the legally relevant facts and disputed issue, and one applying the relevant authorities to explain the reasoning path under the assigned stance.
Closed-book constraints prohibit external search, undisclosed facts, real identities, and placeholder citations; the prompt also hides the authorities actually cited by the court so that the model must select legally responsive authorities on its own.
During de-identification, we remove or mask names, institutions, addresses, and original file paths, and we rewrite the source as a curated background summary before issue extraction.

\subsection{Paired Stance Design}
A paired stance consists of two prompts built from the same factual background and legal issue, where one prompt asks the model to support the court's treatment of the issue and the other asks it to oppose that treatment.
Judgment-prediction systems can learn direct, sometimes legally irrelevant associations between facts and outcomes~\cite{santosh2022deconfounding, medvedeva-mcbride-2023-legal,chen2023knowledgepowerunderstandingcausality}; paired stance prompts are designed to reduce reliance on this shortcut by requiring arguments on both sides of the same issue.
Supporting prompts reward the ability to assemble the facts and authorities that justify the court's result, while opposing prompts reward the ability to identify gaps, missing conditions, or alternative interpretations.
We do not assume that the two stances are equally difficult; the design lets us observe whether a model understands the issue structure rather than only the expected outcome.

\section{Evaluation Protocol}

\ours uses two scoring protocols matched to the structure of each track: reference-answer matching for the public-exam split, and a calibrated three-dimensional rubric for the real-case split.
The following subsections define each scoring scale, the calibration path, and the automatic evaluator.

\begin{table}[t]
\centering
\small
\resizebox{\linewidth}{!}{
\begin{tabular}{p{1.5cm}p{6.5cm}}
\toprule
\textbf{Dimension} & \textbf{Description} \\
\midrule
Citation Relevance & Does the cited authority address the issue, and is the proposition it supports made explicit? \\
\midrule
Constraint Extraction & Does the answer follow the assigned stance, factual boundaries, requested subtasks, and format? \\
\midrule
Argument Validity & Does the conclusion follow from the cited law and facts under the assigned stance? \\
\bottomrule
\end{tabular}
}
\caption{Real-case scoring rubric. Each model answer receives separate integer scores from $0$ to $4$ on the three dimensions. Detailed high-score and low-score behaviors are given in Appendix~\ref{app:rubric-calibration}.}
\label{tab:rubric}
\end{table}

\begin{table}[t]
\centering
\small
\setlength{\tabcolsep}{2pt}
\resizebox{\linewidth}{!}{
\begin{tabular}{lccc}
\toprule
\textbf{Rubric stage} & \textbf{Full avg} & \textbf{3/4 share} & \textbf{Lawyer 1 overlap} \\
\midrule
Initial Checklist & 77.3 & 79.8\% & 76.8 / 75.5 \\
Strict Regrade & 58.0 & 40.0\% & 56.3 / 75.5 \\
Calibrated Real-Case Scores & 73.0 & 70.2\% & 71.3 / 75.5 \\
\bottomrule
\end{tabular}
}
\caption{Three-stage calibration of the real-case rubric. Full avg and 3/4 share use each stage's evaluated pool; the calibrated row covers 276 prompts and 28 models. Lawyer 1 overlap reports automatic mean / Lawyer 1 mean on the same 200 answers across stages.}
\label{tab:calibration}
\end{table}

\begin{table*}[t]
\centering
\small
\setlength{\tabcolsep}{4pt}
\resizebox{\linewidth}{!}{
\begin{tabular}{l*{2}{w{c}{3.3em}}@{\hspace{10pt}}*{5}{w{c}{3.3em}}@{\hspace{10pt}}*{2}{w{c}{3.3em}}}
\toprule
\multirow{2}{*}{\textbf{Model}}
 & \multicolumn{2}{c}{\textbf{Public Exam}}
 & \multicolumn{5}{c}{\textbf{Real Case}}
 & \multicolumn{2}{c}{\textbf{Overall}} \\
\cmidrule(lr){2-3} \cmidrule(lr){4-8} \cmidrule(lr){9-10}
 & \textbf{Auto} & \textbf{Human}
 & \textbf{Cit.} & \textbf{Con.} & \textbf{Arg.} & \textbf{Mean} & \textbf{Human}
 & \textbf{Auto} & \textbf{Human} \\
\midrule
GPT-5.5 High & \textbf{88.0} & \textbf{85.0} & \textbf{85.6} & \textbf{98.7} & \textbf{94.1} & \textbf{92.8} & 77.9 & \textbf{90.4} & 81.5 \\
GPT-5.5 Medium & 79.7 & 75.9 & \underline{81.7} & \underline{98.6} & \underline{92.7} & \underline{91.0} & 79.6 & \underline{85.3} & 77.8 \\
Gemini 3 Flash Low & 85.0 & 80.6 & 74.0 & 74.1 & 71.7 & 73.3 & \underline{85.4} & 79.1 & \textbf{83.0} \\
Qwen3.8-2.4T-A95B & 82.5 & 72.2 & 80.3 & 85.2 & 81.2 & 82.2 & \textbf{87.1} & 82.4 & 79.6 \\
Gemini 3 Flash & 84.6 & \underline{83.1} & 73.8 & 75.6 & 71.3 & 73.6 & 82.5 & 79.1 & \underline{82.8} \\
GPT-5.4 Mini & \underline{85.6} & 75.6 & 73.6 & 90.2 & 84.9 & 82.9 & 76.7 & 84.2 & 76.1 \\
Gemini-3.5-Flash & 85.2 & 79.7 & 73.3 & 74.0 & 72.6 & 73.3 & 81.2 & 79.3 & 80.5 \\
Kimi-K3 & 76.9 & 68.1 & 77.9 & 86.1 & 80.6 & 81.6 & 77.5 & 79.2 & 72.8 \\
GLM-5.3 & 75.8 & 64.7 & 79.1 & 87.0 & 83.2 & 83.1 & 80.0 & 79.5 & 72.3 \\
Grok 4.5 & 74.4 & 74.1 & 74.0 & 89.7 & 81.3 & 81.7 & 72.9 & 78.0 & 73.5 \\
DeepSeek-V4-Pro & 73.7 & 66.9 & 67.2 & 86.1 & 78.9 & 77.4 & 80.0 & 75.6 & 73.4 \\
Kimi K2 Thinking & 76.6 & 69.4 & 72.9 & 75.5 & 72.1 & 73.5 & 76.7 & 75.0 & 73.0 \\
DeepSeek V3.2 & 70.1 & 71.9 & 69.8 & 79.3 & 72.4 & 73.8 & 78.8 & 71.9 & 75.3 \\
DeepSeek R1 & 75.8 & 71.6 & 67.5 & 79.3 & 70.3 & 72.3 & 70.8 & 74.0 & 71.2 \\
Grok 4.3 & 74.4 & 71.9 & 59.1 & 85.3 & 70.1 & 71.5 & 70.8 & 73.0 & 71.4 \\
Gemini 2.5 Flash & 71.9 & 62.5 & 60.3 & 84.6 & 72.0 & 72.3 & 75.4 & 72.1 & 69.0 \\
Deep Cogito 671B & 67.6 & 66.9 & 63.6 & 78.8 & 67.8 & 70.0 & 76.3 & 68.8 & 71.6 \\
GLM-5 low & 68.8 & 66.3 & 68.8 & 81.3 & 72.2 & 74.1 & 70.4 & 71.4 & 68.3 \\
Qwen3 Max Thinking & 68.8 & 62.8 & 70.7 & 82.5 & 74.2 & 75.8 & 71.7 & 72.3 & 67.2 \\
Grok 4.20 & 68.0 & 67.5 & 58.5 & 76.3 & 68.8 & 67.9 & 75.4 & 67.9 & 71.5 \\
Qwen3 Max & 69.5 & 60.9 & 70.2 & 81.8 & 74.5 & 75.5 & 72.1 & 72.5 & 66.5 \\
DeepSeek-V4-Flash & 69.9 & 61.6 & 67.1 & 67.6 & 68.1 & 67.6 & 72.9 & 68.8 & 67.2 \\
Qwen3.8-27B & 67.7 & 58.4 & 68.2 & 80.3 & 71.6 & 73.3 & 72.5 & 70.5 & 65.5 \\
Qwen3 235B A22B & 66.3 & 62.2 & 61.2 & 69.2 & 65.6 & 65.3 & 62.5 & 65.8 & 62.3 \\
Command A & 62.3 & 59.7 & 52.4 & 80.2 & 64.2 & 65.6 & 68.3 & 64.0 & 64.0 \\
Qwen3.5-9B & 59.1 & 45.6 & 61.9 & 69.7 & 65.3 & 65.6 & 75.0 & 62.4 & 60.3 \\
Command R+ & 58.2 & 56.9 & 43.5 & 61.7 & 56.9 & 54.0 & 58.8 & 56.1 & 57.8 \\
LLaMA 3.1 8B & 45.6 & 42.2 & 17.5 & 42.8 & 34.5 & 31.6 & 40.8 & 38.6 & 41.5 \\
\bottomrule
\end{tabular}
}
\caption{Model performance on \ours. Cit./Con./Arg.: citation relevance, constraint extraction, argument validity. Best and second-best values per column are \textbf{bold} and \underline{underlined} (ties included). Models are ranked by the unrounded, equal-weight mean of Overall Auto and Human. Exam scores use reference-answer matching; case scores use the calibrated rubric. Human scores cover all 28 groups: independent reviews of \nexamval exam items and the mean of two lawyers on \nrealcaseval case prompts. \textbf{Overall Auto} averages Exam Auto and Case Mean; \textbf{Overall Human} averages the two Human columns. Overall is descriptive, not an equated ability scale.}
\label{tab:overall}
\end{table*}

\begin{figure*}[t]
\centering
\includegraphics[width=\textwidth]{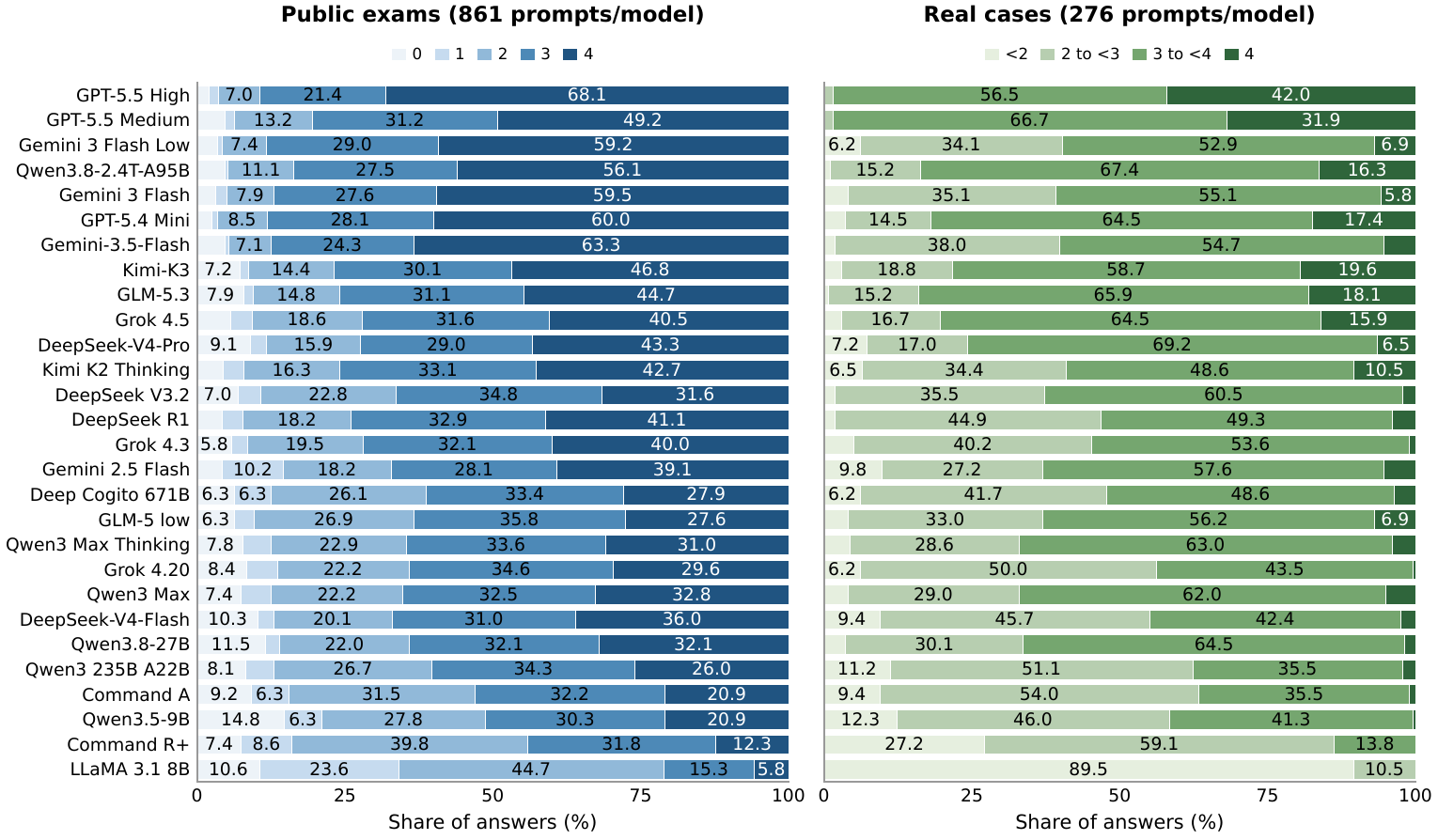}
\caption{Model score distributions for 28 groups. Public-exam bands are integer scores; real-case bands use the per-answer three-dimension mean. Each bar totals 100\% before rounding. In-bar values are percentages to one decimal; narrow segments omit labels. These distributions do not measure repeated-scoring stability.}
\label{fig:score-distribution}
\end{figure*}

\subsection{Public-Exam Scoring}
The public-exam split is scored against the reference answer on an integer scale from $0$ to $4$.
A score of $4$ means that the answer covers the decisive legal points and reaches a compatible conclusion; $3$ means that the core direction is correct but some elements are missing; $2$ means that the answer contains relevant legal content but applies it incompletely; $1$ means only generic or weakly related legal statements; and $0$ means that the answer is wrong, irrelevant, or not evaluable.
For comparison across tracks, all scores are linearly mapped to a $0$--$100$ scale by multiplying by $25$.

\subsection{Real-Case Scoring Rubric}
We adapt CourtReasoner's three-dimensional evaluation framework to the closed-book Chinese case setting~\cite{han2025courtreasoner}.
Citation relevance asks whether the legal authority selected by the answer responds to the issue; constraint extraction assesses compliance with the stance, factual boundaries, and task constraints; argument validity asks whether the conclusion follows from the cited authorities and facts under the assigned stance.
Each dimension receives an integer score from $0$ to $4$, and the real-case score is the arithmetic mean of citation relevance, constraint extraction, and argument validity mapped to the $0$--$100$ range.
\autoref{tab:rubric} summarizes the three dimensions; full high- and low-score behavioral descriptions are in Appendix~\ref{app:rubric-calibration}.

\subsection{Rubric Calibration}
We calibrated the real-case rubric in three stages, summarized in \autoref{tab:calibration}.
An initial checklist rubric was easy to apply but produced inflated scores, with $79.8\%$ of dimension scores at $3$ or $4$ and strong models approaching ceiling performance.
A strict regrading rubric corrected the ceiling but over-penalized ordinary omissions, pushing automatic scores far below human scores on the overlapping subset.
The final calibrated rubric preserves strict caps for nonresponsive, unreadable, or stance-reversed answers, while allowing partial credit for usable legal or factual analysis; the dimension-specific cap rules are given in Appendix~\ref{app:rubric-calibration}.

\subsection{Automatic Evaluator}
\label{sec:auto-evaluator}
The automatic evaluator scores answers against the supplied reference material and rubric without independent legal research. Scoring revisions use Codex~5.5 through the \texttt{gpt-5.5} alias without a pinned model date or explicit sampling parameters, keeping the first valid output after retries. For each instance, the evaluator receives the prompt, the scoring criteria, and one anonymized answer, and returns the $0$--$4$ score(s) of the corresponding track; packet fields and score provenance are in Appendix~\ref{app:scoring-packets}.
Hiding model identity, prior scores, and review notes reduces anchoring and removes explicit model-family cues, but cannot rule out stylistic self-preference, which matters because GPT-5.5 variants are also evaluated, so we validate automatic scores against expert review in \S\ref{sec:analysis-validation}.

\section{Experiments and Results}

\subsection{Experiment Setup}
\label{sec:experiment-setup}

\paragraph{Models.}
We evaluate \nmodel model groups across both tracks, spanning proprietary and open-source families and covering standard and reasoning-mode variants where available. The full model list is provided in Appendix~\ref{app:model-groups}.

\paragraph{Generation.}
Each model answers the prompts under the task-level constraints in \S\ref{sec:benchmark-design}; generation and score provenance are in Appendix~\ref{app:scoring-packets}.

\paragraph{Aggregation and Normalization.}
Public-exam scores are computed over \nexam prompts per model. Real-case scores are computed over \nrealcase prompts per model, with citation relevance, constraint extraction, and argument validity averaged into a real-case mean.
Scores are mapped to a common $0$--$100$ range by multiplying raw values by $25$; this does not equate difficulty across tracks.

\paragraph{Statistical Reporting.}
We use transfer to denote the cross-model association between exam and case scores, reporting Pearson and Spearman correlations over \nmodel model groups. Bootstrap settings are reported in Appendix~\ref{app:bootstrap}.

\subsection{Main Results}
\begin{table}[t]
\centering
\footnotesize
\setlength{\tabcolsep}{4pt}
\renewcommand{\arraystretch}{0.92}
\resizebox{\linewidth}{!}{%
\begin{tabular}{lrrrr}
\toprule
\textbf{Exam group} & \textbf{N} & \textbf{Avg} & \textbf{Pearson} & \textbf{Spearman} \\
\midrule
All public exams & \nexam & 72.6 & 0.817 & 0.708 \\
China & \nexamcn & 60.2 & 0.788 & 0.704 \\
English aggregate & \nexamen & 74.1 & 0.813 & 0.670 \\
United States & \nexamus & 74.0 & 0.820 & 0.677 \\
United Kingdom & \nexamuk & 82.5 & 0.719 & 0.550 \\
Australia & \nexamau & 65.7 & 0.760 & 0.608 \\
\bottomrule
\end{tabular}%
}
\caption{Model-level transfer from public legal exams to Chinese real-case analysis. The real-case endpoint uses the calibrated rubric. English aggregate combines U.S., U.K., and Australian samples.}
\label{tab:country-transfer}
\end{table}

\label{sec:experiment-main}

\begin{tcolorbox}[colback=gray!10, colframe=gray!50!white, width=\linewidth, boxrule=0.5mm, arc=1mm, outer arc=1mm, left=2mm, right=2mm]
\textcolor{orange!50!yellow}{\faLightbulbO}\xspace
\textbf{RQ1:} \rqone
\end{tcolorbox}

\noindent The public-exam mean is $72.6$ and the real-case mean is $73.0$ across the \nmodel model groups (\autoref{tab:overall}).
Citation relevance averages $66.9$, constraint extraction $79.3$, and argument validity $72.6$.
For constraint extraction, $11.9\%$ of answers receive $2$ and $48.9\%$ receive $4$, whereas argument validity has $45.0\%$ at $3$ and $24.7\%$ at $4$. Full dimension distributions appear in \autoref{tab:score-distribution}.
Citation relevance receives lower mean scores than argument validity under both automatic and lawyer evaluation (\S\ref{sec:analysis-dimension}).

\autoref{fig:score-distribution} gives percentages over \nexam exam or \nrealcase case answers per model. Values use one decimal; narrow segments omit labels.

Public-exam scores correlate with Chinese real-case scores at the model level, but the relationship is not deterministic.
\autoref{tab:country-transfer} reports model-level correlations between public-exam means and Chinese real-case means: the full public-exam mean reaches Pearson $r=0.817$ and Spearman $\rho=0.708$, while the Chinese exam subset reaches $r=0.788$ and $\rho=0.704$ despite its lower absolute mean. With only \nmodel model groups, bootstrap intervals for these correlations are wide (Appendix~\ref{app:bootstrap}).
These associations do not isolate language or legal-system effects from differences in source and task composition.
Rankings between the two settings also diverge: top-tier exam models do not always retain their order on the real-case track, and several mid-tier exam models reach competitive real-case scores (\autoref{tab:overall}, \autoref{fig:transfer-main}; per-model labels and lawyer-scored results are in Appendix~\ref{app:transfer-scatterplots}).

\begin{figure}[t]
\centering
\begin{tikzpicture}[lbl/.style={font=\tiny, inner sep=1pt}]
\begin{axis}[
    width=\columnwidth,
    height=5.0cm,
    xmin=40, xmax=95, ymin=25, ymax=100,
    xtick={40,50,60,70,80,90}, ytick={30,40,50,60,70,80,90,100},
    xlabel={Public-exam score (0--100)},
    ylabel={Real-case score (0--100)},
    tick label style={font=\scriptsize},
    label style={font=\scriptsize},
    axis line style={draw=gray!50},
    tick style={draw=gray!50},
    grid=major, grid style={gray!15},
]
\addplot[gray!70, dashed, domain=40:95, samples=2] {x};
\addplot[only marks, mark=*, mark size=1.5pt, draw=blue!70, fill=blue!35] coordinates {(88.0,92.8) (79.7,91.0) (85.0,73.3) (82.5,82.2) (84.6,73.6) (85.6,82.9) (85.2,73.3) (76.9,81.6) (75.8,83.1) (74.4,81.7) (73.7,77.4) (76.6,73.5) (70.1,73.8) (75.8,72.3) (74.4,71.5) (71.9,72.3) (67.6,70.0) (68.8,74.1) (68.8,75.8) (68.0,67.9) (69.5,75.5) (69.9,67.6) (67.7,73.3) (66.3,65.3) (62.3,65.6) (59.1,65.6) (58.2,54.0) (45.6,31.6)};
\addplot[only marks, mark=*, mark size=1.5pt, draw=orange!80!black, fill=orange!60] coordinates {(88.0,92.8) (79.7,91.0) (75.8,83.1) (85.2,73.3) (85.0,73.3) (84.6,73.6) (45.6,31.6)};
\node[lbl, anchor=south] at (axis cs:88.0,94.0) {GPT-5.5 High};
\node[lbl, anchor=east] at (axis cs:78.7,91.0) {GPT-5.5 Medium};
\node[lbl, anchor=south] at (axis cs:75.8,84.3) {GLM-5.3};
\node[lbl, anchor=north] at (axis cs:85.0,72.0) {Gemini 3/3.5 Flash};
\node[lbl, anchor=west] at (axis cs:46.6,31.6) {LLaMA 3.1 8B};
\node[anchor=north west, font=\scriptsize, align=left, fill=white, draw=gray!40, inner sep=2pt] at (rel axis cs:0.03,0.97) {Pearson $r=0.817$\\ Spearman $\rho=0.708$};
\end{axis}
\end{tikzpicture}
\caption{Model-level exam-to-case transfer under automatic scoring for the \nmodel model groups. The dashed line marks equal normalized scores and is a scale reference only. Per-model labels and the lawyer-scored counterpart are in Appendix~\ref{app:transfer-scatterplots}.}
\label{fig:transfer-main}
\end{figure}
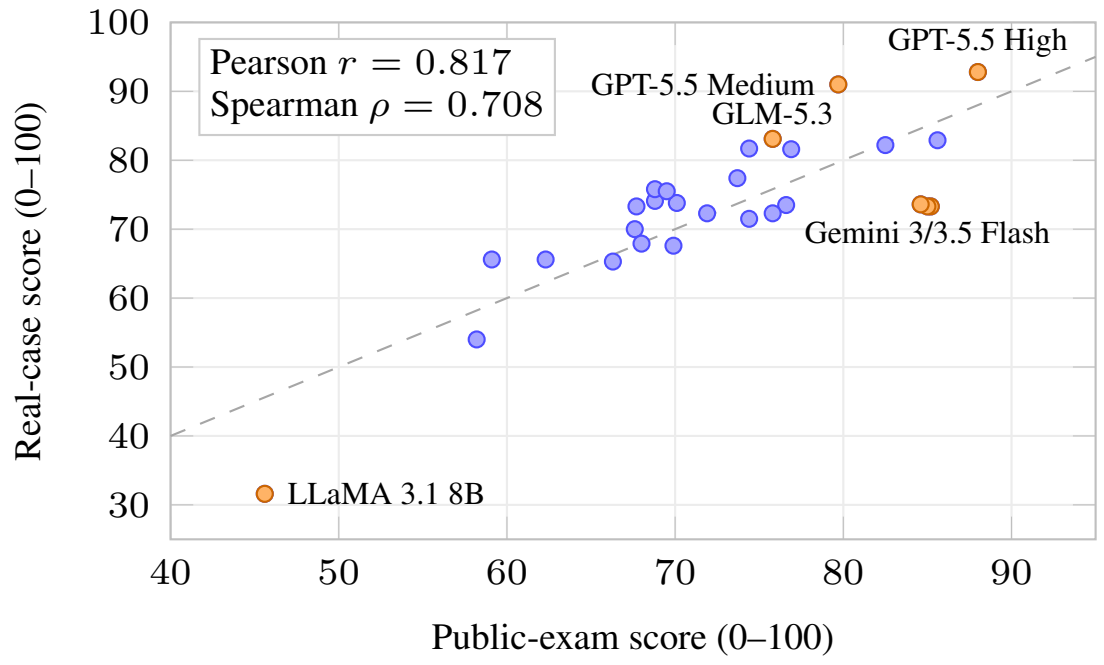

We assess the citation--argument contrast through prompt- and judgment-cluster resampling and independent lawyer scores. Five scoring repetitions per dimension measure evaluator variability on 200 fixed answers. Domain, language, and C-Eval controls examine the exam--case association under different exam subsets and model coverage~\citep{huang2023cevalmultilevelmultidisciplinechinese}. Sample sizes, scoring conditions, and results appear in Appendix~\ref{app:bootstrap}.

\section{Analysis}
\vspace{-0.015cm}
\begin{tcolorbox}[colback=gray!10, colframe=gray!50!white, width=\linewidth, boxrule=0.5mm, arc=1mm, outer arc=1mm, left=2mm, right=2mm]
\textcolor{orange!50!yellow}{\faLightbulbO}\xspace
\textbf{RQ2:} \rqtwo
\end{tcolorbox}

\noindent We probe the main results through four analyses of where and how real-case legal reasoning fails (\S\ref{sec:analysis-dimension}--\S\ref{sec:analysis-error}), and then ask whether automated scoring agrees with expert review (\S\ref{sec:analysis-validation}).

\subsection{Dimension-Level Analysis}
\label{sec:analysis-dimension}

The dimension gap in \S\ref{sec:experiment-main} concerns citation relevance: models receive lower mean scores for selecting responsive authorities than for constructing arguments under a given stance.
Citation relevance and argument validity assess different aspects of a response, so this contrast does not establish an intrinsic difficulty hierarchy.
\autoref{tab:realcase-category} shows lower mean citation relevance than argument validity in all seven case categories. Tort is the largest category (29.0\%), while medical-malpractice prompts make up 18.1\% of the case split.

\begin{table}[t]
\centering
\footnotesize
\setlength{\tabcolsep}{4pt}
\begin{tabular}{lrrrrr}
\toprule
\textbf{Category} & \textbf{N} & \textbf{Cit.} & \textbf{Con.} & \textbf{Arg.} & \textbf{Avg} \\
\midrule
Tort & 80 & 67.7 & 75.7 & 72.4 & 71.9 \\
Contract & 72 & 66.0 & 81.2 & 73.1 & 73.4 \\
Criminal & 54 & 68.4 & 81.1 & 72.6 & 74.0 \\
Intellectual Property & 34 & 67.1 & 80.8 & 73.3 & 73.7 \\
Administrative & 22 & 62.5 & 80.0 & 70.3 & 70.9 \\
Civil Procedure & 8 & 63.3 & 77.2 & 70.5 & 70.3 \\
Property & 6 & 74.4 & 82.4 & 76.2 & 77.7 \\
\bottomrule
\end{tabular}
\caption{Chinese real-case performance by category under automatic scoring. N counts issue-stance prompts; each prompt is answered by \nmodel model groups. Cit./Con./Arg. as in \autoref{tab:overall}.}
\label{tab:realcase-category}
\end{table}

\subsection{Reasoning-Mode and Family Analysis}
\label{sec:analysis-family}

We next ask how reasoning-oriented or larger model variants differ across tracks.
\autoref{tab:family-effects} reports six comparisons across five families.
Variant differences do not transfer uniformly to real-case analysis.
GPT-5.5 High scores $+2.4$ points above GPT-5.4 Mini on the public exam and $+9.9$ on the real-case mean; relative to GPT-5.5 Medium it gains $+8.3$ on the exam but only $+1.8$ on the real-case mean.
Qwen3 Max Thinking gains $+0.5$ on citation relevance but loses $0.3$ on argument validity, while DeepSeek R1 gains $+5.7$ on the exam but loses $1.5$ on the real-case mean.
These contrasts include differences in checkpoint and model size as well as reasoning mode, and therefore do not isolate the effect of reasoning effort.

\begin{table}[t]
\centering
\small
\setlength{\tabcolsep}{2pt}
\resizebox{\linewidth}{!}{
\begin{tabular}{lrrrr}
\toprule
\textbf{Comparison} & \textbf{Exam $\Delta$} & \textbf{Cit. $\Delta$} & \textbf{Con. $\Delta$} & \textbf{Case $\Delta$} \\
\midrule
GPT-5.5 High - GPT-5.4 Mini & +2.4 & +12.0 & +8.5 & +9.9 \\
GPT-5.5 High - GPT-5.5 Medium & +8.3 & +3.9 & +0.2 & +1.8 \\
Gemini 3 Flash - Low & -0.3 & -0.2 & +1.5 & +0.3 \\
Qwen3 Max Think - Non-Think & -0.7 & +0.5 & +0.7 & +0.3 \\
DeepSeek R1 - DeepSeek V3.2 & +5.7 & -2.4 & +0.0 & -1.5 \\
Grok 4.5 - Grok 4.3 & 0.0 & +14.9 & +4.3 & +10.2 \\
\bottomrule
\end{tabular}
}
\caption{Six comparisons across five model families. Checkpoint, size, and reasoning-mode differences are not isolated. Positive values mean the first model scores higher. Cit./Con. as in \autoref{tab:overall}.}
\label{tab:family-effects}
\end{table}

\subsection{Source and Domain Analysis}
\label{sec:analysis-source}

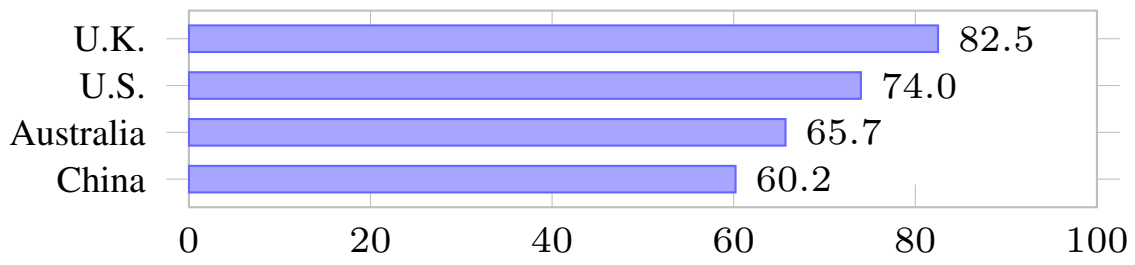
\begin{figure}[t]
\centering
\begin{tikzpicture}
\begin{axis}[
    xbar,
    scale only axis,
    width=0.78\columnwidth,
    height=1.3cm,
    xmin=0, xmax=100,
    enlarge y limits=0.2,
    bar width=5pt,
    symbolic y coords={China, Australia, U.S., U.K.},
    ytick=data,
    nodes near coords={\pgfmathprintnumber[fixed,fixed zerofill,precision=1]{\pgfplotspointmeta}},
    nodes near coords align={horizontal},
    every node near coord/.append style={font=\scriptsize},
    tick label style={font=\scriptsize},
    axis line style={draw=gray!50},
    tick style={draw=gray!50},
]
\addplot[fill=blue!35, draw=blue!60] coordinates {(60.2,China) (65.7,Australia) (74.0,U.S.) (82.5,U.K.)};
\end{axis}
\end{tikzpicture}
\caption{Public-exam means by country jurisdiction across \nmodel model groups. U.S.\ subsource and topic-category breakdowns are reported in Appendix~\ref{app:source-domain-breakdowns}.}
\label{fig:source-effects}
\end{figure}

We next examine how public-exam performance varies across country jurisdictions, U.S.\ subsources, and legal categories; these differences should be read as source and format effects rather than intrinsic jurisdictional difficulty.
\autoref{fig:source-effects} summarizes country-level means; the full U.S.\ source and topic breakdowns are reported in Appendix~\ref{app:source-domain-breakdowns}.
U.K.\ samples score highest at $82.5$, followed by U.S.\ at $74.0$, Australia at $65.7$, and China at $60.2$; U.S.\ subsources also vary, including fictional State of Columbia items that require following prompt-specific rather than real-world authority.

\vspace{-0.03cm}

\subsection{Error Analysis}
\label{sec:analysis-error}
Real-case scoring notes reveal five recurring failure mechanisms (\autoref{tab:failures-taxonomy}).
Their common thread is not a lack of legal vocabulary: answers cite plausible rules in fluent prose, yet the conclusion rests on an over-broad rule, a missing condition, or facts added beyond the prompt.
RV038 (\autoref{fig:teaser}) is typical: answers supporting the refusal of re-appraisal assert appraisal details absent from the closed record, keeping most citation and argument credit while losing constraint extraction.
Error cards and the full case study are in Appendix~\ref{app:error-cards}.

\begin{table}[t]
\centering
\fontsize{8.5}{10}\selectfont
\setlength{\tabcolsep}{3pt}
\renewcommand{\arraystretch}{1.1}
\begin{tabularx}{\columnwidth}{@{}>{\raggedright\arraybackslash}p{0.225\columnwidth}>{\raggedright\arraybackslash}X@{}}
\toprule
\textbf{Mechanism} & \textbf{Typical behavior} (\textit{dimensions affected}) \\
\midrule
Citation mismatch & Cited authority is irrelevant, outdated, or too broad for the proposition (\textit{Cit.; often Arg.}). \\
Condition omission & Omits operative conditions such as fault, causation, or burden of proof (\textit{Con.; often Arg.}). \\
Fact extrapolation & Adds expert opinions, evidence, or other facts absent from the prompt (\textit{Con., Arg.}). \\
Stance drift & Argues against the assigned stance or shifts it midway (\textit{Con., Arg.; severe cap}). \\
Conclusion jump & Moves from an abstract rule to a conclusion without fact mapping (\textit{Arg.}). \\
\bottomrule
\end{tabularx}
\caption{Failure mechanisms in \S\ref{sec:analysis-error} and the rubric dimensions they most affect. Cit./Con./Arg.: citation relevance, constraint extraction, argument validity.}
\label{tab:failures-taxonomy}
\end{table}

\subsection{Human Validation}
\label{sec:analysis-validation}
\begin{tcolorbox}[colback=gray!10, colframe=gray!50!white, width=\linewidth, boxrule=0.5mm, arc=1mm, outer arc=1mm, left=2mm, right=2mm]
\textcolor{orange!50!yellow}{\faLightbulbO}\xspace
\textbf{RQ3:} \rqthree
\end{tcolorbox}

\noindent The public-exam validation subset contains \nexamval exam items answered by 28 model groups (\nexamvalans answers scored by independent reviewers against the reference answers); the real-case subset contains \nrealcaseval randomly sampled issue-stance prompts (\nrealcasevalans answers from the same 28 groups).
Two practicing Chinese lawyers, both excluded from benchmark construction, independently scored the same randomized answer set with model identities, automatic scores, and each other's ratings hidden; Lawyer 2's ratings were held out from rubric calibration.
\autoref{tab:human-validation} reports automatic--human agreement, using the equal-weight mean of the two lawyers as the real-case Human score. Exam scoring aligns strongly with human review, whereas real-case agreement is lower at the answer level ($r=.312$, $\rho=.235$) than after model-level aggregation ($r=.703$, $\rho=.459$).

The two lawyers' quadratic-weighted kappa is $.676/.560/.559$ for citation relevance, constraint extraction, and argument validity, and their model-level means correlate at $r=.841$ ($\rho=.711$). Both give citation relevance a lower mean than argument validity (pooled difference $5.13$ points), but the prompt-cluster 95\% intervals for the pooled and Lawyer 2 scores cross zero, so the difference is descriptive, not statistically significant, and we claim no consistent ordering for constraint extraction.
\begin{table}[t]
\centering
\small
\setlength{\tabcolsep}{4pt}
\begin{tabular}{lrrr}
\toprule
\textbf{Validation set}  & \textbf{Model} & \textbf{Human} & \textbf{Answer-level} \\
\midrule
Public exams & 71.2 & 67.3 & \(r=.910, \rho=.898\) \\
Real cases & 71.7 & 73.9 & \(r=.312, \rho=.235\) \\
\bottomrule
\end{tabular}
\caption{Automatic scoring versus human validation. Model-level correlations are \(r=.948, \rho=.963\) for public exams and \(r=.703, \rho=.459\) for real cases.}
\label{tab:human-validation}
\end{table}

\section{Conclusion}

We introduced \ours, pairing public legal exams with lawyer-reviewed issue-stance prompts from de-identified Chinese judgments. Exam scores track case performance at the model level but remain a coarse proxy: rankings and reasoning-mode gains shift across tracks, so exam leaderboards alone should not guide model selection for case-grounded legal work. Citation relevance receives lower mean scores than argument validity under automatic and lawyer evaluation. Because real-case failures stem less from missing legal knowledge than from weakly responsive authorities and reasoning beyond the record, progress likely requires better authority selection and record-faithful argumentation, assessed with expert review wherever automatic judges diverge at the answer level. The benchmark complements exams with expert-grounded case diagnostics and measures exam--case associations and differences between automatic and expert evaluation.

\section*{Limitations}

First, the real-case split covers \nrealcase prompts from \njudgment Chinese judgments with uneven category coverage, so the findings may not generalize to other legal systems or case types.
Future work should add practitioner-reviewed cases from other jurisdictions and languages, prioritizing under-represented categories.
Second, the prompts are curated closed-book records rather than full case files, so they do not test long-dossier skills such as reconciling conflicting evidence across procedural documents.
Pairing exam items and case prompts on the same doctrine, and adding full-dossier tasks where privacy permits, would further separate application failures from topic and format effects.
Third, public-exam items may appear in pretraining data, so exam scores are comparative signals rather than clean generalization tests.
Future releases should add contamination-controlled exam pools, such as held-out items collected after a model freeze date.
Finally, scoring updates rely on a single LLM evaluator whose model family is also evaluated, and lawyer double annotation covers only \nrealcaseval prompts (\nrealcasevalans answers).
Broader double annotation and an independent evaluator are needed before automatic scores can replace expert review~\cite{10.1093/jla/laae003, magesh2024hallucinationfreeassessingreliabilityleading}.

\section*{Ethics and Data Governance}

Real-case prompts are built from Chinese judgments that were de-identified before issue extraction (\S\ref{sec:benchmark-design}); models are never asked to predict real litigants or recover identities, and external search is prohibited.
\ours is a diagnostic research benchmark, not a legal-advice system: a high score reflects performance under a controlled protocol, not competence to advise clients, decide cases, or replace legal professionals.
Our results show that fluent legal analysis can still rest on the wrong authority or on facts outside the record, a risk that applies to any deployment.

\paragraph{Data and code release.}
We release all \nrealcase case prompts with scoring references, the \nrealcaseans case responses with individual scoring records, model metadata, and evaluation code (links below the abstract).
To respect third-party copyright, the exam release is limited to 78 Victorian Bar excerpts under CC BY-NC-ND 4.0; the original judgments, the remaining 783 exam inputs, and exam response text are not released.

\bibliography{refs}

\clearpage
\appendix
\raggedbottom

\section{Evaluation Setup}
\phantomsection
\subsection{Model Groups}
\label{app:model-groups}
The benchmark evaluates \nmodel model groups: GPT-5.5 High, GPT-5.4 Mini, GPT-5.5 Medium, Gemini 3 Flash, Gemini 3 Flash Low, Gemini 2.5 Flash, Kimi K2 Thinking, DeepSeek R1, DeepSeek V3.2, Grok 4.5, Grok 4.3, Grok 4.20, GLM-5 low, Qwen3 Max Thinking, Qwen3 Max, Qwen3 235B A22B, Deep Cogito 671B, Command A, Command R+, LLaMA 3.1 8B Instruct, Qwen3.5-9B, Qwen3.8-27B, Qwen3.8-2.4T-A95B, GLM-5.3, Kimi-K3, DeepSeek-V4-Pro, DeepSeek-V4-Flash, and Gemini-3.5-Flash. Human validation covers all 28 groups. Display names are harmonized across tracks; original call identifiers are retained in the records.

\subsection{Real-Case Scoring Packet Fields}
\label{app:scoring-packets}

Each real-case scoring packet contains one model answer, the prompt, citation basis, cited proposition, review constraints, and rubric. The assigned stance and core issue appear in the prompt or as separate fields. The scorer does not receive the model name, previous dimension scores, or previous review notes. This packet design is intended to reduce anchoring across model answers and prevent explicit model-family labels from affecting the evaluation.

\paragraph{Evaluator configuration and retries.}
Scoring revisions use Codex 5.5 through the \texttt{gpt-5.5} alias, without a pinned model date or explicit \texttt{temperature}, \texttt{top\_p}, or random seed. Each revised score is the first valid result after execution, parsing, or schema retries; repeated judgments are not averaged into the headline scores. Repeatability is assessed in isolated scoring sessions; evaluation configurations are recorded in the scoring logs.

\paragraph{Score provenance.}
Reported exam responses and scores predate the prompt and reference-text repairs. For real-case answers, constraint extraction was rescored under the English-compliant rubric in Appendix~\ref{app:rubric-calibration}; enforcing the B=$0$ cap also required two C-score corrections. Other scores were retained where no rescoring was performed. Review notes were not regenerated with B and therefore do not explain the revised B scores. Lawyer ratings were retained unchanged, so comparisons with B also reflect differences in rubric version and language.

\subsection{Prompt and Scoring Packet Example}
\label{app:packet-example}

\autoref{tab:packet-example} shows a compressed example of the information visible to the evaluator. The evaluator receives one model answer at a time, together with the case fields needed for scoring. The evaluator does not receive the model name, previous dimension scores, or previous review notes.

\begin{table*}[t]
\centering
\small
\begin{tabularx}{\textwidth}{p{0.20\textwidth}X}
\toprule
\textbf{Packet field} & \textbf{Compressed example from RV038} \\
\midrule
Issue & Whether the court should permit re-appraisal when the appraisal opinion is alleged to be unfair. \\
Position & Supports the court judgment. \\
Core issue & The court did not grant the plaintiff's application for re-appraisal and still used the existing appraisal materials as the basis for judgment. \\
Citation basis & Article 40 of the Several Provisions of the Supreme People's Court on Evidence in Civil Proceedings. \\
Cited proposition & Re-appraisal requires statutory conditions such as serious procedural illegality or an appraisal opinion that clearly lacks sufficient basis; dissatisfaction with the conclusion alone is not enough. \\
Review constraints & Do not add off-topic facts, undisplayed evidence, unrecorded diagnosis or treatment details, external search results, real names, or institutional identities. \\
Answer visibility & A single model answer is shown without model identity or old scoring information. \\
\bottomrule
\end{tabularx}
\caption{Compressed example of a real-case scoring packet.}
\label{tab:packet-example}
\end{table*}

\subsection{Example of Model Input and Output}
\label{app:model-input-output}

This appendix shows representative input-output structures for the two benchmark tracks.
\autoref{tab:cn-judgment-example} shows a Chinese real-case issue-stance prompt, which uses the closed-book setting and the three-dimensional calibrated rubric.
\autoref{tab:bar-exam-example} shows a public-exam prompt and its scoring target, which is scored against a reference answer.

\begin{table*}[t]
\centering
\small
\setlength{\tabcolsep}{6pt}
\renewcommand{\arraystretch}{1.08}
\begin{tabularx}{\textwidth}{|X|}
\hline
\textbf{Issue title.} Re-appraisal in a medical-malpractice dispute. \\
\hline
\textbf{Source body.} De-identified Chinese civil judgment involving medical-damage liability and an existing appraisal opinion. The original judgment was converted into a closed-book issue-stance prompt after removal or masking of names, institutions, addresses, identification numbers, contact information, monetary identifiers, and original file paths. \\
\hline
\textbf{Stance.} Support the court result. \\
\hline
\textbf{Prompt excerpt.} Write a two-paragraph Chinese legal analysis supporting the court's refusal to grant re-appraisal. Use only the supplied facts. Do not add undisclosed medical records, appraisal defects, treatment details, institutional identities, or external case information. The first paragraph should summarize the legally relevant facts and disputed issue. The second paragraph should apply the relevant legal authorities to explain why the court's handling is defensible. \\
\hline
\textbf{Evaluator reference.} The court did not grant the plaintiff's application for re-appraisal and relied on the existing appraisal materials as the basis for judgment. Dissatisfaction with the appraisal conclusion alone is not enough; re-appraisal requires statutory or rule-based grounds such as serious procedural illegality or an appraisal opinion that clearly lacks sufficient basis. \\
\hline
\textbf{Evaluator citation basis.} Article 40 of the Several Provisions of the Supreme People's Court on Evidence in Civil Proceedings; relevant medical-damage liability rules concerning fault, causation, appraisal opinions, and liability apportionment. \\
\hline
\textbf{Review constraints.} Closed-book setting. Do not use external search. Do not invent expert opinions, diagnosis details, medical records, institutional identities, undisclosed evidence, or facts not shown in the prompt. Preserve the assigned stance. The evaluator sees only one model answer at a time and does not see the model name, old scores, or old review notes. \\
\hline
\textbf{Model answer excerpt (Qwen3 235B A22B).} In this case, the original appraisal was made by an appraisal institution with statutory qualifications after lawfully accepting the commission; during the appraisal process, all outpatient and emergency medical records, electrocardiogram materials, and emergency treatment records were reviewed, and the compliance of the diagnosis and treatment conduct, the sufficiency of the risk disclosure, and causation were analyzed item by item, with complete procedures and sufficient basis. \\
\hline
\textbf{Rubric consequence.} A strong answer should identify the legal threshold for re-appraisal, connect that threshold to the displayed appraisal posture, and avoid adding undisclosed appraisal defects or medical facts. In weaker answers, models cite relevant appraisal rules but supplement the closed-book record with unstated defects or stronger causation facts, lowering the constraint-extraction score even when citation relevance or argument validity remains moderate or high. \\
\hline
\end{tabularx}
\caption{Compressed RV038 prompt with a verbatim excerpt from the stored English model answer (A/B/C: 4/1/3). The excerpt asserts appraisal details not supplied in the prompt. The evaluator-facing packet appears in \autoref{tab:packet-example}.}
\label{tab:cn-judgment-example}
\end{table*}

\begin{table*}[t]
\centering
\small
\setlength{\tabcolsep}{6pt}
\renewcommand{\arraystretch}{1.08}
\begin{tabularx}{\textwidth}{|X|}
\hline
\textbf{Issue title.} Permission to appeal from the Court of Appeal (Civil Division). \\
\hline
\textbf{Source body.} Solicitors Regulation Authority, SQE1 FLK1 sample question, England and Wales. Dataset ID: \texttt{EXTUKSQE1FLK1Q10}. Unlike the Chinese real-case split, this track is scored against a reference answer rather than the three-dimensional real-case rubric. \\
\hline
\textbf{Stance.} Not stance-paired. The model should answer the examination question directly. \\
\hline
\textbf{Prompt facts.} A decision is made by the Court of Appeal (Civil Division) in favour of the claimant. The defendant wishes to obtain permission to appeal. \\
\hline
\textbf{Question.} Which of the following courts have the power to grant permission to appeal? A. The Court of Appeal only. B. The Supreme Court only. C. The House of Lords only. D. The Court of Appeal and the Supreme Court. E. The Court of Appeal and the House of Lords. \\
\hline
\textbf{Review constraints.} Answer only the official examination item represented by this row. First state the best answer choice, then briefly explain why it is correct. Base the answer only on the reproduced facts/materials in this prompt. Do not add outside facts. \\
\hline
\textbf{Reference answer.} D. The Court of Appeal and the Supreme Court. \\
\hline
\textbf{Model output excerpt (GLM-5.3).} Best answer: D. The Court of Appeal and the Supreme Court. Explanation: Following a decision of the Court of Appeal (Civil Division), permission to appeal to the Supreme Court may be granted either by the Court of Appeal itself or by the Supreme Court. \\
\hline
\textbf{Scoring target.} Substantive consistency with the reference answer. The recorded automatic score is 4/4. Although the source item supplies answer choices, the model also produces an explanation. \\
\hline
\end{tabularx}
\caption{A public-exam item and an excerpt from its stored model response. Text is reproduced from the evaluation packet; formatting is simplified.}
\label{tab:bar-exam-example}
\end{table*}

\subsection{Rubric Calibration Details}
\label{app:rubric-calibration}

\autoref{tab:rubric-detail} gives the full high-score and low-score behaviors used when applying the three real-case dimensions.
\autoref{tab:appendix-calibration} then summarizes the three-stage calibration path: the initial checklist rubric was useful for implementation but produced a ceiling effect, the strict regrade corrected the ceiling but over-penalized ordinary omissions, and the calibrated rubric maintains a high standard for score $4$ while removing unnecessary global caps.
Specifically, the calibrated rubric does not treat ordinary omissions as non-responses, sets citation relevance to $0$ when no legal authority is given while allowing constraint extraction and argument validity to retain limited credit when substantive analysis is present, and caps argument validity at $2$ rather than forcing it to $0$ when constraint extraction is $0$.
Constraint extraction assesses compliance with the prompt's stance, factual boundaries, requested subtasks, and format; English answers satisfy the language requirement. A missing identifiable authority caps B at $2$, a wrong stance gives B=$0$, and serious fabrication caps B at $1$. Score provenance is documented in Appendix~\ref{app:scoring-packets}.

\begin{table*}[t]
\centering
\small
\begin{tabularx}{\textwidth}{p{0.14\textwidth} X X}
\toprule
\textbf{Dimension} & \textbf{High-score behavior} & \textbf{Low-score behavior} \\
\midrule
Citation relevance & Identifies relevant statutes, judicial interpretations, or legal rules and explains the proposition they support. & Gives no legal authority, cites irrelevant or unsupported authority, or lists authorities without explaining their connection to the dispute. \\
Constraint extraction & Preserves the assigned stance and the supplied factual and evidentiary boundaries, addresses the requested subtasks, and follows paragraph, language, and privacy requirements. English answers satisfy the language requirement. & Omits required subtasks, ignores fact boundaries, adds unstated facts, reverses the stance, or violates format and privacy constraints. \\
Argument validity & Maps rule conditions to facts, handles uncertainty, and forms a legally defensible reason under the assigned stance. & Jumps from abstract rules to conclusion, extrapolates facts, breaks the reasoning chain, reverses the stance, or reaches a conclusion incompatible with the cited authority and facts. \\
\bottomrule
\end{tabularx}
\caption{Detailed high-score and low-score behaviors for each real-case rubric dimension. This expands the summary in \autoref{tab:rubric}.}
\label{tab:rubric-detail}
\end{table*}

\begin{table*}[t]
\centering
\small
\begin{tabularx}{\textwidth}{lrrlX}
\toprule
\textbf{Rubric stage} & \textbf{Full avg} & \textbf{3/4 share} & \textbf{Lawyer 1 overlap} & \textbf{Scoring behavior} \\
\midrule
Initial checklist & 77.3 & 79.8\% & 76.8 / 75.5 & Too permissive; many fluent answers received high scores even when they missed key constraints. \\
Strict regrade & 58.0 & 40.0\% & 56.3 / 75.5 & Too harsh; ordinary omissions were sometimes treated like severe non-responses. \\
Calibrated rubric & 73.0 & 70.2\% & 71.3 / 75.5 & Partial credit under dimension-specific caps. \\
\bottomrule
\end{tabularx}
\caption{Rubric calibration for the Chinese real-case split. Full avg and 3/4 share use each stage's evaluated pool; the calibrated row covers 276 prompts and 28 models. Lawyer 1 overlap reports automatic mean / Lawyer 1 mean on the same 200 answers across stages.}
\label{tab:appendix-calibration}
\end{table*}

\section{More Statistics}
\subsection{Score Distribution by Dimension}
\label{app:score-distribution}

\autoref{tab:score-distribution} reports the detailed score distribution
for each real-case dimension, complementing the dimension-level results
reported in \S\ref{sec:experiment-main}.

\begin{table}[t]
\centering
\small
\begin{tabular}{lrrrrr}
\toprule
\textbf{Dimension} & \textbf{0} & \textbf{1} & \textbf{2} & \textbf{3} & \textbf{4} \\
\midrule
Citation (A) & 141 & 399 & 2412 & 3641 & 1135 \\
Constraint (B) & 145 & 543 & 921 & 2337 & 3782 \\
Argument (C) & 12 & 278 & 2053 & 3480 & 1905 \\
\bottomrule
\end{tabular}
\caption{Score counts by dimension over \nrealcaseans Chinese real-case answers.}
\label{tab:score-distribution}
\end{table}

\begin{table}[t]
\centering
\small
\begin{tabular}{lrr}
\toprule
\textbf{Score pattern} & \textbf{Count} & \textbf{Share} \\
\midrule
3/4/3 & 1009 & 13.1\% \\
3/4/4 & 885 & 11.5\% \\
3/3/3 & 810 & 10.5\% \\
4/4/4 & 654 & 8.5\% \\
2/4/3 & 528 & 6.8\% \\
2/3/3 & 454 & 5.9\% \\
2/3/2 & 424 & 5.5\% \\
2/4/2 & 269 & 3.5\% \\
\bottomrule
\end{tabular}
\caption{Most frequent score combinations in the Chinese real-case split. Each pattern lists citation relevance, constraint extraction, and argument validity scores in that order.}
\label{tab:score-combinations}
\end{table}

\subsection{Transfer Scatterplots}
\label{app:transfer-scatterplots}

\begin{figure*}[t]
\centering
\begin{subfigure}{0.98\textwidth}
    \centering
    \includegraphics[width=\linewidth]{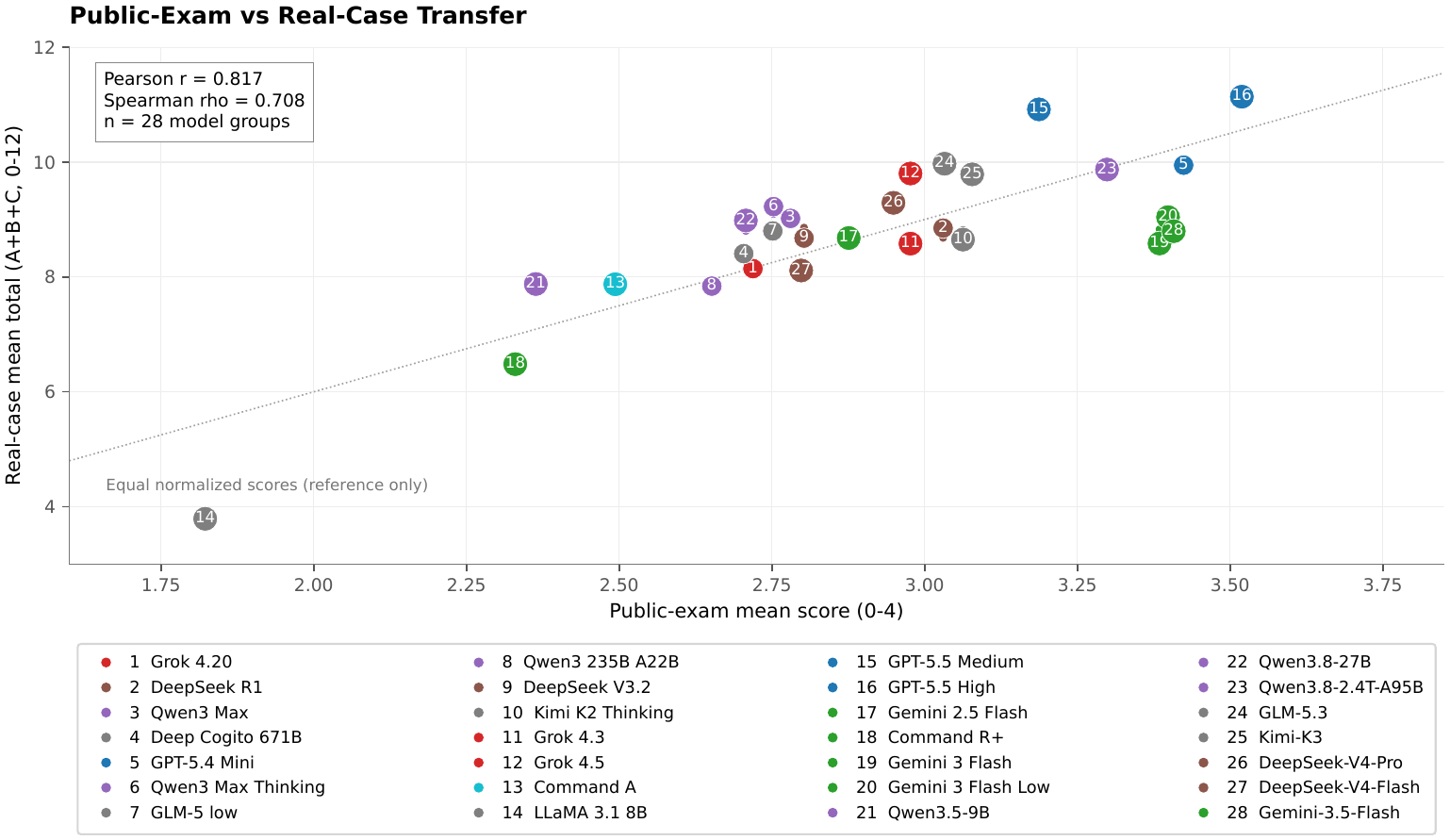}
    \caption{Model-judge exam-to-case transfer.}
    \label{fig:model-judge-transfer-appendix}
\end{subfigure}

\vspace{0.6em}

\begin{subfigure}{0.98\textwidth}
    \centering
    \includegraphics[width=\linewidth]{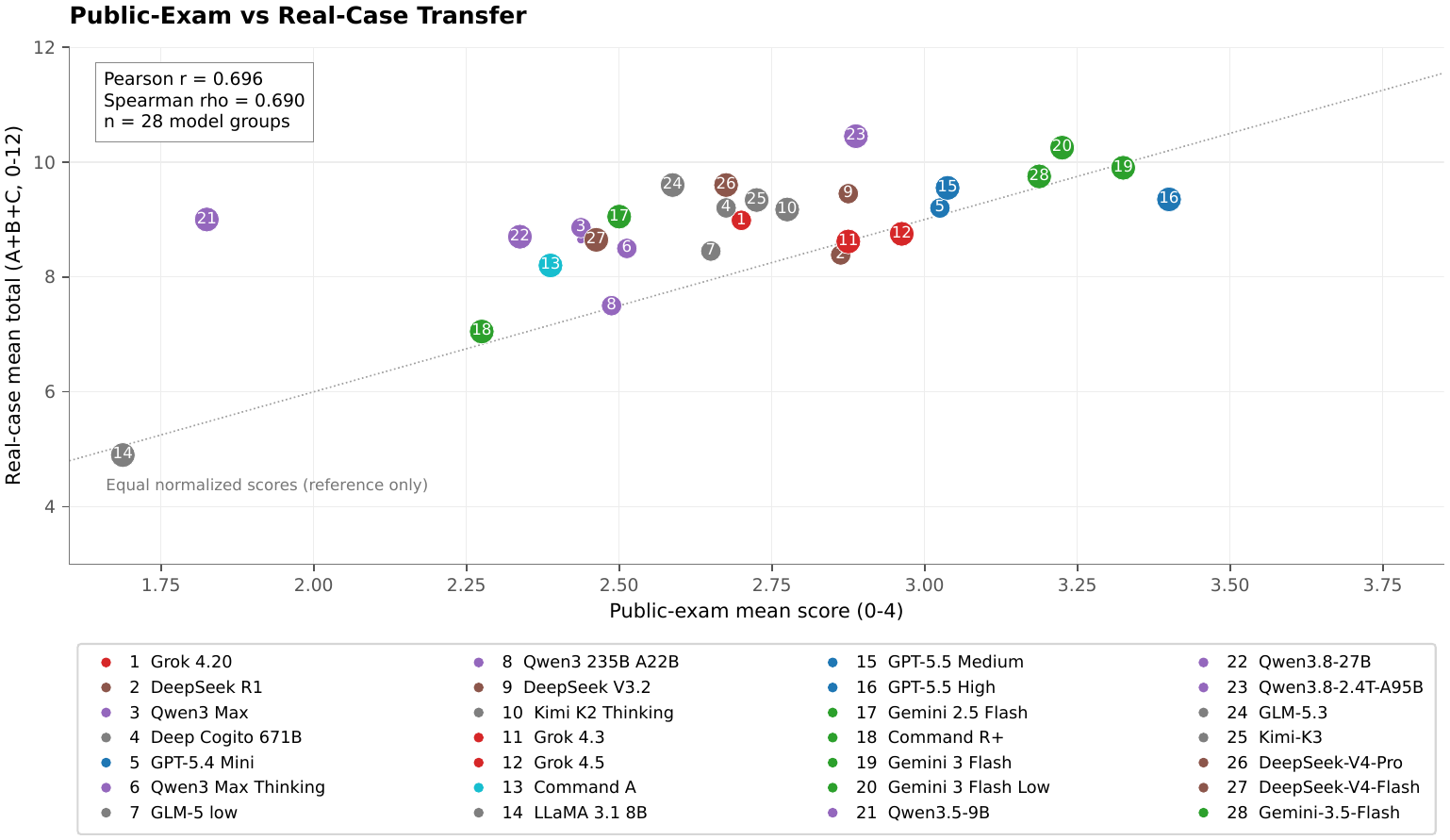}
    \caption{Human exam scores and pooled lawyer case scores.}
    \label{fig:human-transfer-appendix}
\end{subfigure}

\caption{Exam-to-case associations. Each point represents one model group: 28 in each panel. Panel (a) uses automatic scores; panel (b) uses human exam scores and the equal-weight mean of both lawyers' case scores on the validation subsets, not automatic--human agreement. The diagonal marks equal normalized scores, not equated task difficulty.}
\label{fig:transfer-scatterplots}
\end{figure*}

\autoref{fig:transfer-scatterplots} visualizes the model-level exam--case association. Panel (a) uses automatic scores for 28 model groups. Panel (b) uses human exam scores and pooled lawyer case scores for 28 groups on the validation subsets. Neither panel measures automatic--human agreement, and the diagonal is a scale reference rather than an ability-loss threshold.

\subsection{Source and Domain Breakdowns}
\label{app:source-domain-breakdowns}

This appendix reports the full source- and topic-level breakdowns that contextualize the country-level summary in \autoref{fig:source-effects}.
\autoref{tab:us-source} shows mean scores across U.S.\ subsources, including UBE materials, NCBE MEE reprints, and state-level bar exams.
\autoref{tab:exam-topics} reports public-exam category means for categories with at least \ntopicmin samples.

\begin{table}[t]
\centering
\small
\resizebox{\linewidth}{!}{
\begin{tabular}{lrr}
\toprule
\textbf{U.S. source / legal-system label} & \textbf{N} & \textbf{Avg} \\
\midrule
New York / NCBE MEE reprint & 102 & 77.4 \\
Maryland UBE law & 75 & 77.0 \\
California general U.S. law & 100 & 74.0 \\
Uniform Bar Examination law & 57 & 73.8 \\
Virginia / U.S. bar law & 182 & 73.2 \\
Georgia bar exam law & 70 & 70.4 \\
California fictional State of Columbia & 17 & 64.3 \\
\bottomrule
\end{tabular}
}
\caption{U.S. public-exam source differences. Means are model-answer averages on the 0--100 scale. Small sources should not be over-interpreted.}
\label{tab:us-source}
\end{table}

\begin{table}[t]
\centering
\small
\begin{tabularx}{\linewidth}{Xrr}
\toprule
\textbf{Legal category} & \textbf{N} & \textbf{Avg} \\
\midrule
Foundations of Legal Knowledge & 68 & 83.8 \\
Business Organizations / Agency & 126 & 75.0 \\
Secured Transactions / Commercial Law & 67 & 74.5 \\
Tort & 242 & 73.7 \\
Family & 58 & 72.1 \\
Professional Responsibility / Legal Ethics & 38 & 71.0 \\
Succession / Trusts / Estates & 61 & 70.2 \\
Civil Procedure & 26 & 68.2 \\
Civil / Civil Procedure & 22 & 62.5 \\
Criminal Procedure & 23 & 56.1 \\
Commercial / Company Law & 23 & 55.7 \\
\bottomrule
\end{tabularx}
\caption{Public-exam performance by legal category. The table includes categories with at least \ntopicmin samples.}
\label{tab:exam-topics}
\end{table}

\subsection{Prompt-Level Score Variation}
\label{app:case-difficulty}

\autoref{tab:case-difficulty} reports prompt-level score variation over all \nrealcase review IDs. Means and sample standard deviations are computed across the \nmodel model answers for each prompt after averaging the three dimensions and converting to the 0--100 range. These statistics describe differences among models on a prompt, not variation between raters or complete judgments.

\begin{table*}[t]
\centering
\small
\resizebox{\textwidth}{!}{
\begin{tabular}{rllllll}
\toprule
\textbf{Rank} & \textbf{Lowest-avg prompt} & \textbf{Avg/SD} & \textbf{Highest-avg prompt} & \textbf{Avg/SD} & \textbf{Largest cross-model SD} & \textbf{Avg/SD} \\
\midrule
1 & RV024 & 52.1/17.5 & RV184 & 92.0/16.3 & RV115 & 63.7/24.3 \\
2 & RV119 & 52.1/16.5 & RV231 & 90.5/12.2 & RV245 & 67.3/22.0 \\
3 & RV159 & 52.4/16.3 & RV168 & 89.9/11.6 & RV057 & 67.6/21.4 \\
4 & RV189 & 53.6/14.1 & RV180 & 89.6/10.3 & RV061 & 67.0/21.2 \\
5 & RV127 & 53.9/16.4 & RV112 & 89.3/11.5 & RV062 & 76.8/21.2 \\
6 & RV145 & 54.2/17.5 & RV164 & 88.4/13.5 & RV265 & 59.5/21.1 \\
7 & RV147 & 54.8/14.6 & RV088 & 88.1/12.3 & RV072 & 75.3/21.1 \\
8 & RV161 & 55.1/17.9 & RV128 & 87.8/9.8 & RV050 & 80.1/21.1 \\
9 & RV232 & 55.1/14.8 & RV178 & 87.8/9.5 & RV001 & 70.5/20.8 \\
10 & RV273 & 55.1/14.6 & RV262 & 87.8/14.3 & RV070 & 68.2/20.8 \\
\bottomrule
\end{tabular}
}
\caption{Prompt-level score variation in the Chinese real-case split. Each mean and standard deviation is computed across \nmodel model answers for the same review ID, over all 276 prompts. SD describes variation across models, not disagreement between raters.}
\label{tab:case-difficulty}
\end{table*}

\subsection{Human-Model Disagreement}
\label{app:human-disagreement}

\autoref{tab:human-disagreement} lists representative model-level differences on the \nrealcasevalans overlapping answers, using pooled lawyer scores. Rank difference is automatic rank minus pooled-lawyer rank among all 28 models, with average ranks for ties. Positive values indicate a higher rank under lawyer scoring. Differences in either direction show that aggregate means do not establish answer-level agreement or identify the cause of a scoring difference.

\begin{table*}[t]
\centering
\small
\begin{tabular}{lrrrr}
\toprule
\textbf{Model} & \textbf{Automatic} & \textbf{Pooled lawyers} & \textbf{Gap} & \textbf{Rank diff.} \\
\midrule
Qwen3 235B A22B & 71.7 & 62.5 & +9.2 & -10.0 \\
Deep Cogito 671B & 67.5 & 76.3 & -8.8 & +8.0 \\
Gemini 3 Flash Low & 74.2 & 85.4 & -11.2 & +9.0 \\
Qwen3 Max & 75.8 & 72.1 & +3.8 & -12.5 \\
Gemini 3 Flash & 70.8 & 82.5 & -11.7 & +15.0 \\
GPT-5.4 Mini & 91.7 & 76.7 & +15.0 & -10.0 \\
Qwen3 Max Thinking & 80.0 & 71.7 & +8.3 & -15.0 \\
\bottomrule
\end{tabular}
\caption{Representative model-level differences on the same 280 answers. Pooled lawyers is the equal-weight mean of both lawyers; Gap is Automatic minus Pooled lawyers. Rank differences use all 28 models and average ranks for ties.}
\label{tab:human-disagreement}
\end{table*}

\subsection{Robustness Audits}
\label{app:bootstrap}

For exam-to-case transfer, we resample the \nmodel model groups with replacement 10,000 times (NumPy PCG64, seed 20260712); \autoref{tab:bootstrap-transfer} reports model-level percentile intervals. These are conditional on the evaluated roster and do not account for dependence within model families. Dimension contrasts use separate clustered resampling.

\begin{table*}[t]
\centering
\small
\begin{tabular}{lrrrr}
\toprule
\textbf{Exam group} & \textbf{Pearson} & \textbf{95\% CI} & \textbf{Spearman} & \textbf{95\% CI} \\
\midrule
All exams & 0.817 & [0.563, 0.928] & 0.708 & [0.401, 0.888] \\
China & 0.788 & [0.543, 0.900] & 0.704 & [0.404, 0.868] \\
U.S. & 0.820 & [0.558, 0.933] & 0.677 & [0.347, 0.878] \\
U.K. & 0.719 & [0.434, 0.860] & 0.550 & [0.200, 0.794] \\
Australia & 0.760 & [0.499, 0.876] & 0.608 & [0.277, 0.821] \\
\bottomrule
\end{tabular}
\caption{Model-level 95\% percentile intervals from 10,000 draws over 28 model groups. These describe the evaluated roster and do not account for dependence within model families.}
\label{tab:bootstrap-transfer}
\end{table*}

\paragraph{Domain, language, and general-Chinese controls.}
The supplementary controls below use a case endpoint distinct from the main 276-prompt, 28-model evaluation. The Tort control compares 243 exam items with 76 case prompts from 15 judgments across 20 model configurations, giving $r=.807$ and $\rho=.554$; its 116 medical-malpractice and 127 other-Tort subsets give $r=.815/\rho=.568$ and $r=.792/\rho=.573$. The English-exam aggregate in the language control gives $r=.820/\rho=.636$, compared with $r=.868/\rho=.805$ for the Chinese subset. On 14 groups mappable to public C-Eval scores~\citep{huang2023cevalmultilevelmultidisciplinechinese}, C-Eval gives $r=.808/\rho=.074$, whereas the Chinese legal exam gives $r=.831/\rho=.558$; removing LLaMA~3.1~8B changes the former Pearson coefficient to $-.522$ and the latter to $.616$. A separate sensitivity using 10 nearby-version mappings gives $r=-.331/\rho=-.579$ for C-Eval, $r=-.327/\rho=-.455$ for CMMLU~\citep{li2024cmmlumeasuringmassivemultitask}, and $r=.873/\rho=.863$ for the Chinese legal exam. These exploratory mappings are not matched-checkpoint tests or evidence that language is fully controlled.

For argument validity minus citation relevance ($C-A$), we resample prompts or judgments 10,000 times over all 276 prompts from 56 judgments (7,728 answers). A sampled prompt retains all 28 model answers; a sampled judgment retains all its prompts. We also give each judgment equal weight in a sensitivity analysis. Human resampling retains all 28 models and both lawyers within each of the 10 sampled prompt clusters. We use NumPy PCG64 with seed 20260712 and 95\% percentile intervals. \autoref{tab:clustered-robustness} shows positive point estimates throughout, although the pooled and Lawyer 2 intervals cross zero. A separate cluster audit covers 76 prompts and 20 models.

\begin{table*}[t]
\centering
\small
\setlength{\tabcolsep}{5pt}
\begin{tabular}{llrr}
\toprule
\textbf{Scoring source} & \textbf{Resampling unit} & \textbf{$C-A$} & \textbf{95\% CI} \\
\midrule
Automatic evaluator & Prompt clusters & 5.69 & [4.51, 6.83] \\
Automatic evaluator & Judgment clusters & 5.69 & [4.44, 6.92] \\
Automatic evaluator & Judgments (equal weight) & 6.02 & [4.84, 7.17] \\
Lawyer 1 & Prompt clusters & 8.12 & [2.59, 13.93] \\
Lawyer 2 & Prompt clusters & 2.14 & [-5.18, 8.93] \\
Pooled lawyers & Prompt clusters & 5.13 & [-0.94, 11.07] \\
\bottomrule
\end{tabular}
\caption{Argument validity minus citation relevance ($C-A$), with 10,000-draw percentile intervals. Automatic rows use 276 prompts, 56 judgments, and 28 models (7,728 answers); lawyer rows use the same 10 prompts and 28 models (280 answers). Each draw retains all responses within a sampled prompt or judgment; the pooled analysis retains both lawyers.}
\label{tab:clustered-robustness}
\end{table*}

To assess evaluator stability, we repeat scoring five times for each of the three dimensions on the same 200 blinded answers from 10 prompts and 20 model configurations. We report exact agreement, the proportion of answers whose scores span at most one point, mean pairwise quadratic weighted kappa, and mean per-answer standard deviation. Across repetitions, mean scores range from $69.00$--$70.13$ for citation relevance, $67.13$--$70.63$ for constraint extraction, and $73.00$--$74.63$ for argument validity. Tables~\ref{tab:evaluator-stability}--\ref{tab:evaluator-round-means} summarize consistency and dimension means. For constraint extraction, 10,000 prompt-cluster draws give exact-agreement CI $[39.0,51.0]\%$, range-at-most-one CI $[69.5,81.5]\%$, QWK CI $[.737,.846]$, and mean answer-SD CI $[.320,.424]$. These analyses quantify scoring variability for fixed answers; the headline results use the recorded single scores.

\begin{table*}[t]
\centering
\small
\setlength{\tabcolsep}{5pt}
\begin{tabular}{lrrrr}
\toprule
\textbf{Dimension} & \textbf{Five-run exact} & \textbf{Range $\leq 1$} & \textbf{Mean pairwise QWK} & \textbf{Mean answer SD} \\
\midrule
Citation relevance & 57.5\% & 99.5\% & 0.867 & 0.213 \\
Constraint extraction & 45.0\% & 75.5\% & 0.802 & 0.372 \\
Argument validity & 67.5\% & 99.5\% & 0.764 & 0.163 \\
\bottomrule
\end{tabular}
\caption{Scoring consistency over five repetitions per dimension on 200 fixed answers.}
\label{tab:evaluator-stability}
\end{table*}

\begin{table*}[t]
\centering
\small
\setlength{\tabcolsep}{7pt}
\begin{tabular}{lrrrrr}
\toprule
 & \multicolumn{5}{c}{\textbf{Mean score (0--100)}} \\
\cmidrule(lr){2-6}
\textbf{Dimension} & \textbf{Run 1} & \textbf{Run 2} & \textbf{Run 3} & \textbf{Run 4} & \textbf{Run 5} \\
\midrule
Citation relevance (A) & 69.125 & 69.125 & 70.125 & 70.000 & 69.000 \\
Constraint extraction (B) & 67.125 & 70.125 & 69.375 & 70.625 & 69.750 \\
Argument validity (C) & 74.625 & 73.625 & 73.000 & 74.375 & 74.250 \\
Argument minus citation ($C-A$) & 5.500 & 4.500 & 2.875 & 4.375 & 5.250 \\
\bottomrule
\end{tabular}
\caption{Scoring consistency over five repetitions per dimension on 200 fixed answers.}
\label{tab:evaluator-round-means}
\end{table*}

\section{Error Analysis}
\subsection{Qualitative Error Cards}
\label{app:error-cards}

\autoref{tab:error-cards} summarizes representative error cards for the five failure modes in \autoref{tab:failures-taxonomy} (\S\ref{sec:analysis-error}), selected from the scoring notes. These examples are anonymized at the case-fact level and are intended to explain error mechanisms, not to provide a complete qualitative audit of any single model.

\begin{table*}[t]
\centering
\small
\begin{tabularx}{\textwidth}{>{\raggedright\arraybackslash}p{0.15\textwidth}>{\raggedright\arraybackslash}p{0.08\textwidth}>{\raggedright\arraybackslash}p{0.15\textwidth}>{\raggedright\arraybackslash}p{0.08\textwidth}>{\raggedright\arraybackslash}X}
\toprule
\textbf{Failure mechanism} & \textbf{Review ID} & \textbf{Model} & \textbf{Scores} & \textbf{Short explanation} \\
\midrule
Citation mismatch & RV001 & Command R+ & 1/0/1 & The answer relies on generic or outdated contract authority and adds an unsupported after-sales-service prerequisite, so the cited law does not support the assigned opposition argument. \\
Condition omission & RV063 & Gemini 3 Flash Low & 4/2/3 & The answer gives a legally plausible damages discussion but does not control causation, successor liability, liability ratio, and damage-item boundaries required by the prompt. \\
Fact extrapolation & RV038 & Qwen3 235B A22B & 4/1/3 & The answer cites relevant appraisal authority and gives usable re-appraisal reasoning, but it asserts detailed appraisal findings and procedures not present in the closed-book record. \\
Stance drift & RV001 & Command R+ & 1/0/1 & The response begins from the assigned opposition posture but shifts toward a conclusion that does not squarely oppose the court's handling outcome. \\
Conclusion jump & RV063 & Gemini 3 Flash Low & 4/2/3 & The answer moves from general damages rules to a result without sufficiently mapping the prompt's causation and liability-share constraints. \\
No usable legal basis & RV011 & LLaMA 3.1 8B Instruct & 0/0/0 & The answer gives no usable statutory or judicial-interpretation basis and contains severe garbling and unstable factual phrasing, so the calibrated rubric assigns all-zero scores. \\
\bottomrule
\end{tabularx}
\caption{Representative qualitative error cards from the Chinese real-case scoring notes, grouped by failure mechanism.}
\label{tab:error-cards}
\end{table*}

\paragraph{Case Study: RV038.}
RV038 illustrates a constraint-extraction failure.
The issue asks whether a court should permit re-appraisal in a medical-malpractice dispute after an existing appraisal had been issued and challenged by the plaintiff.
Several model answers identified relevant medical-liability rules and produced coherent arguments about appraisal review, yet the main failure was not a lack of legal vocabulary; the weaker answers supplemented the closed-book record with assumed appraisal defects, unstated medical findings, or stronger causal claims than the prompt provided.
Under the calibrated rubric, this pattern can receive substantial credit for citation relevance or argument form while still losing points on constraint extraction because the answer does not preserve the factual and evidentiary boundaries of the case.
\autoref{tab:case-study-rv038} gives the structured mechanism breakdown, showing how a response can look legally plausible while failing the real-case requirement that legal reasoning remain tied to the supplied record.

\begin{table*}[!t]
\centering
\small
\setlength{\tabcolsep}{5pt}
\renewcommand{\arraystretch}{0.95}

\begin{tabularx}{\textwidth}{p{0.17\textwidth}X}
\toprule
\textbf{Element} & \textbf{RV038 mechanism} \\
\midrule

Case issue &
Whether re-appraisal should be permitted in a medical-malpractice dispute after an
existing appraisal had been issued and challenged by the plaintiff. \\

Expected constraint &
The answer should reason from the displayed appraisal posture and avoid adding
undisclosed defects, medical records, or alternative expert findings. \\

Observed behavior &
Some answers cite relevant medical-liability or appraisal rules and give a
plausible legal path, but infer extra appraisal deficiencies or stronger causation
facts not stated in the prompt. \\

Rubric consequence &
Citation relevance and argument validity can remain moderate or high, but constraint
extraction is reduced because the answer crosses the closed factual and evidentiary
boundary. \\

\bottomrule
\end{tabularx}

\caption{Case study of the low-constraint-extraction/high-argument-validity
pattern in RV038. The case illustrates the difference between fluent legal
argumentation and controlled constraint extraction.}
\label{tab:case-study-rv038}
\end{table*}

\end{document}